\documentclass{article}

\usepackage{microtype}
\usepackage{graphicx}
\usepackage{subfigure}
\usepackage{booktabs} 
\usepackage{amsmath}
\usepackage{hyperref}

\PassOptionsToPackage{table}{xcolor}
\usepackage[accepted]{icml2024}

\usepackage{amssymb}
\usepackage{mathtools}
\usepackage{amsthm}

\usepackage[capitalize,noabbrev]{cleveref}

\theoremstyle{plain}

\theoremstyle{definition}

\theoremstyle{remark}

\usepackage[textsize=tiny]{todonotes}

\usepackage{multicol}
\usepackage{multirow}
\usepackage{tabularx}
\newcolumntype{Y}{>{\centering\arraybackslash}X}
\newcolumntype{Z}{>{\centering\arraybackslash\hsize=1.5\hsize}X}

\begin{document}

\twocolumn[
\icmltitle{SKIM: Any-bit Quantization Pushing The Limits of Post-Training Quantization}




\begin{icmlauthorlist}
\icmlauthor{Runsheng Bai}{thu}
\icmlauthor{Bo Liu}{uta}
\icmlauthor{Qiang Liu}{uta}
\end{icmlauthorlist}

\icmlaffiliation{thu}{School of Software, Tsinghua University, Beijing, China}
\icmlaffiliation{uta}{Department of Computer Science, University of Texas at Austin}

\icmlcorrespondingauthor{Qiang Liu}{lqiang@cs.utexas.edu}


\vskip 0.3in
]




\begin{abstract}
Large Language Models (LLMs) exhibit impressive performance across various tasks, but deploying them for inference poses challenges. Their high resource demands often necessitate complex, costly multi-GPU pipelines, or the use of smaller, less capable models. While quantization offers a promising solution utilizing lower precision for model storage, existing methods frequently experience significant performance drops at lower precision levels. Additionally, they typically provide only a limited set of solutions at specific bit levels, many of which are extensively manually tuned. To address these challenges, we propose a new method called \textbf{SKIM}: Scaled K-means clustering wIth Mixed precision. Our approach introduces two novel techniques: 1. A \textit{greedy algorithm} to solve approximately optimal bit allocation across weight channels, and 2. A \textit{trainable scaling vector} for non-differentiable K-means clustering. These techniques substantially improve performance and can be adapted to any given bit. Notably, in terms of model perplexity, our method narrows the gap between 3-bit quantized LLaMA models and their full precision counterparts by \textbf{16.3\%} on average.
\end{abstract}

\section{Introduction}

Large Language Models (LLMs) including GPT \cite{radford2019language} and LLaMA \cite{touvron2023llama}, have achieved remarkable performance across a diverse range of tasks. These models not only excel in language processing \cite{achiam2023gpt, dubey2024llama, chowdhery2023palm, zhang2022opt} but also adapt effectively to multimodal applications \cite{wang2024visionllm, driess2023palm}, marking a crucial step toward artificial general intelligence \cite{bubeck2023sparks}. However, the computational and memory demands of LLMs pose significant challenges. For instance, when loading parameters in FP16, GPT requires 350GB of memory, while LLaMA-65B needs at least 130GB, both of which far exceeding the capabilities of A100-80G GPUs. Even when conducting inference with the smallest LLaMA model, which has 7 billion parameters, an Out-of-Memory exception can occur on a widely used 24GB GPU. These challenges significantly complicate the storage and practical deployment of such models.

\begin{figure}[t]
  \centering
  \includegraphics[width=\columnwidth]{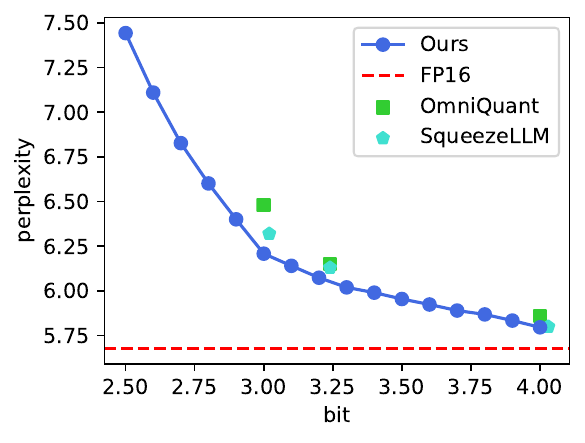}
  \label{fig:intro}
  \vspace{-20pt}
  \caption{Our SKIM method adaptively quantizes the model to any specified bit and achieves superior performance. The perplexity reported is of LLaMA-7B on the WikiText2 dataset.}
  \vspace{-10pt}
\end{figure}

One promising technique to address these issues is quantization, which involves transforming high-precision data into lower-precision formats, such as converting FP16 parameters to INT4. This method directly reduces the memory required to deploy and load LLMs and improves inference latency due to the phenomenon of the memory wall \cite{gholami2024ai}, which identifies the memory bandwidth as a key bottleneck in LLM inference. In addition, quantization has shown promising performance benefits. For example, previous studies have shown that both LLM weights and activations can be stored in 8 bits \cite{xiao2023smoothquant}, or only LLM weights can be stored in 4 bits \cite{kim2023squeezellm}, with little performance degradation. This encourages researchers to explore lower-precision solutions while maintaining reasonable performance levels. 

However, standard quantization techniques in recent methods can suffer a significant drop in performance when using low bit widths. To mitigate this decline, these methods often introduce additional techniques that incur extra memory costs. For example, SqueezeLLM \cite{kim2023squeezellm} retains certain sensitive elements and outliers with full precision using a sparse tensor, while AWQ \cite{lin2024awq} divides the quantization group into smaller ones, requiring the storage of more quantization factors. Additionally, the extra memory needed to achieve a reasonable trade-off between memory usage and performance often requires manual tuning and selection, making the process cumbersome.

\textbf{Contribution} In this paper, we address the above issues with our proposed method, Scaled K-means clustering wIth Mixed Precision (SKIM), which optimizes the bit allocation using a greedy algorithm and regularizes the column-wise difference with a scaling vector. Our method can easily adapt to any specified bit, including even non-integer values, and achieve better performance. Figure \ref{fig:intro} illustrates how our method breaks the fixed bit grid and delivers better results. Our main contributions can be summarized as follows: (1) We conduct a mathematical analysis of two optimization targets: layer-wise and sensitivity-based quantization, identifying a unified framework that highlights their core differences and allows us to evaluate their effectiveness. (2) We observe a significant disparity in data distribution across channels and propose a greedy algorithm for approximately optimal bit allocation in response to this disparity. Our mixed precision method adapts to any specified bit level and significantly improves performance. (3) For the non-differentiable K-means clustering operator, we incorporate a trainable scaling vector based on our novel iterative optimization strategy. This vector effectively regularizes the data across columns and serves as a valuable complement to the mixed precision method.

\section{Related Work}
\paragraph{Quantization of LLMs} Quantization can be viewed from different perspectives. First, based on whether the method involves training the entire model with quantization in mind, it can be categorized into two types: \textit{Quantization-Aware Training (QAT)} \cite{jacob2018quantization, kim2023full, liu2023llm} and \textit{Post-Training Quantization (PTQ)} \cite{cai2020zeroq, shomron2021post}. Although QAT methods often perform better, their significant resource requirements for retraining the entire model make them less practical for LLMs. Therefore, PTQ is more widely adopted for language models, and also the focus of our work. Additionally, methods can also be classified by whether they both weights and activations are quantized, leading to two categories: \textit{Weight-Activation Quantization} \cite{yao2022zeroquant, xiao2023smoothquant, zhao2024atom} and \textit{Weight-Only Quantization} \cite{frantar2022gptq, heo2023rethinking, lin2024awq}. In this paper, we concentrate on the Weight-Only method, as it has emerged as a promising technique for addressing memory demands and improving inference efficiency.

\paragraph{Non-uniform Method} Non-uniform quantization \cite{jeon2022mr, liu2022nonuniform}, which utilizes varying quantization intervals, is a powerful approach due to its higher proximity to the data distribution. Among various techniques, such as space transformation \cite{yvinec2023nupes} and trainable quantization factors \cite{jeon2022mr}, K-means clustering \cite{krishna1999genetic, kanungo2002efficient, ahmed2020k} is a widely adopted one \cite{xu2018deep, zadeh2020gobo, kim2023squeezellm} for its efficiency and effectiveness. K-means clustering generates cluster labels and centroids, allowing us to store the labels directly as low-bit data and the centroids as a codebook for recovery. Furthermore, weights can be incorporated into K-means clustering to address different optimization targets; for instance, SqueezeLLM \cite{kim2023squeezellm} introduces sensitivity as weights to enhance performance. Our work builds upon these previous efforts and refines them with our own observations and techniques.

\paragraph{Quantization Techniques} Outliers have been a significant obstacle for LLM quantization in achieving lossless solution \cite{dettmers2022gpt3, xiao2023smoothquant}. Previous works have proposed various methods to tackle this issue, including ideas that are similar to our mixed precision technique and scaling vector to some extent \cite{dettmers2023spqr, xiao2023smoothquant}. However, our approach differs remarkably from these prior methods. In terms of mixed precision, previous techniques primarily focus on mixing a specific bit level with the original level, such as INT3 and FP16, and usually apply element-wise mixed precision aiming to maintain the integrity of some crucial elements. In contrast, our method adaptively blends all available bit levels and employs a channel-wise mixture, allowing for better resource allocation. Regarding scaling factors, most existing methods applied it under the uniform quantization, where only differential operators are considered. Our work shifts this focus to a non-uniform context, optimizing it on the non-differential grouping operator with a novel strategy.

\section{Review The Quantization Objectives}
 Previous works have proposed different quantization objectives, significantly broadening the scope of this field. However, these objectives are often considered in isolation, making the evaluation and selection of objectives unnecessarily complicated, especially for our multi-process SKIM method. In this section, we conduct a comparative analysis on two widely adopted approaches: the layer-wise and sensitivity-based objectives. This analysis highlights their similarities and key distinctions, providing foundations for informed selection, which will be discussed in Section \ref{sec:obj}.

\subsection{Notations}
We define a general linear layer using following notations: \\
\vspace{-20pt} 
\begin{itemize}
    \setlength{\leftskip}{-8pt}   
    \setlength{\itemsep}{2pt}
    \item $W \in \mathbb{R}^{n \times m}$ and $X \in \mathbb{R}^{m \times k}$ denote the weight and input matrix, respectively. And the corresponding output matrix is $Y = WX \in \mathbb{R}^{n \times k}$.
    \item The quantized weight, denoted as $W^q \in \mathbb{R}^{n \times m}$, is the full-precision matrix reconstructed from its low-bit representation. It satisfies the constraint that its values are restricted to $2^{bit}$ discrete centroids for each group.
    \item The $i$-th row of $W$ is represented as $w_i \in \mathbb{R}^{1 \times m}$. The same definition applies to $w^q_i$, $x_i$ and $y_i$.
    \item The gradient and Hessian matrix are computed with respect to the final loss $L$. Taking $w_i$ as example, $g_{w_i} = \nabla_{w_i} L \in \mathbb{R}^{1 \times m}$ represents the gradient, and $H_{w_i} = \nabla_{w_i}^2 L \in \mathbb{R}^{m \times m}$ represents the Hessian matrix.
\end{itemize}

\subsection{Layer-wise Quantization}
The Layer-wise Quantization Framework has been widely adopted \cite{frantar2022gptq, hubara2021accurate} to make the task more targeted. This framework aims to quantize each layer individually and addresses corresponding reconstruction problems. Concretely, let $W$ be the full-precision weight matrix, and $X$ the input data. The goal is to find the quantized weight $W^q$ that minimizes the layer-wise squared error between the outputs of the original and quantized weights, which can be formally expressed as:
\begin{equation}
\mathop{\arg\min}\limits_{W^q} \left\| WX - W^qX \right\|^2. \label{eq:lw} 
\end{equation}

\subsection{Sensitivity-based Quantization} 
Instead of minimizing the layer-wise squared error, SqueezeLLM \cite{kim2023squeezellm} proposes minimizing the overall perturbation with respect to the final loss. They use the second-order Taylor expansion to analyze how changes in a specific layer weight $W$ influence the final loss, and further assume the first-order term is approximately zero since the model to be quantized should have already converged. For simplicity in understanding, here we use $w_i$, the $i$-th row of $W$, to explain the target. With its Hessian matrix $H_{w_i}$, the objective can be written as:
\begin{equation}
\mathop{\arg\min}\limits\limits_{w^q_i}\ (w_i-w^q_i)H_{w_i}(w_i-w^q_i)^\top. \label{eq:sb}
\end{equation} 
Furthermore, two additional approximations are incorporated: 1. Take Fisher information matrix computed on a calibration dataset $D$ as a Hessian approximation to avoid heavy computation. It can be formally expressed as $H \approx F = \mathbb{E}(g^\top g) \approx \frac{1}{|D|}\sum_{d\in D} (g^{(d)})^\top g^{(d)}$, where $F$ is the Fisher information matrix and $g^{(d)}$ represents the gradient computed on sample $d$; 2. The Fisher information matrix is further approximated as a diagonal matrix by assuming that cross-weight interactions are negligible, reducing the complexity from quadratic to linear. Consequently, with $\mathrm{diag}(\cdot)$ representing the diagonal function, the final objective can be written as:
\begin{align}
    & \mathop{\arg\min}\limits\limits_{w^q_i}\ (w_i-w^q_i)\ \mathrm{diag}(\mathcal{F}_{w_i})\ (w_i-w^q_i)^\top \\  
    =\ & \mathop{\arg\min}\limits\limits_{w^q} \sum_{j=1}^m {(g_{w_i})_j}^2\cdot(w_{ij}-w^q_{ij})^2. \label{eq:sb-diag}
\end{align}

\begin{figure*}[tb]
  \centering
  \includegraphics[width=\textwidth]{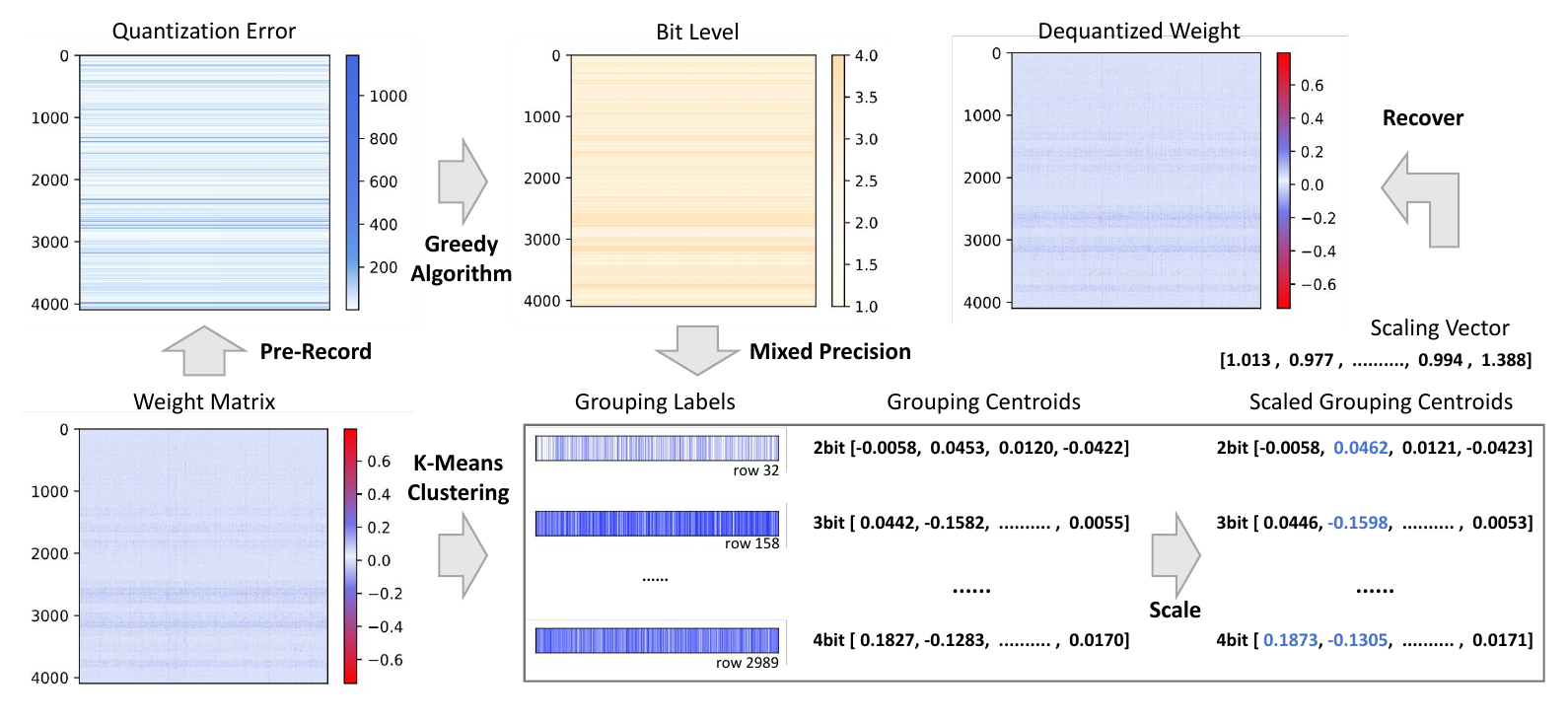}
  \vspace{-20pt}
  \caption{Overall procedure of our proposed SKIM algorithm. The method consists of three main part: greedy algorithm for bit allocation, weighted K-Means Clustering based on allocation, and the trainable scaling vector. More details are available in Section \ref{sec:method}.
  }
  \label{fig:process}
\end{figure*}

\subsection{Reformulation}
Although layer-wise and sensitivity-based objectives have different assumptions and initial forms, the following reformulation aims to organize them into unified structures, either \textit{full} or \textit{diag}, which will be discussed later. This analysis will provide a foundation for our subsequent work.

\paragraph{Layer-wise quantization} Since the L2-norm of a matrix can be expressed as the sum of that of its rows, Equation (\ref{eq:lw}) can be reformulated in terms of the $i$-th row of $W$:
\begin{align}
    & \mathop{\arg\min}\limits\limits_{w^q_i} \left\| (w_i - w^q_i)X \right\|^2 \label{eq:lf} \\ 
    =\ & \mathop{\arg\min}\limits\limits_{w^q_i}\ (w_i - w^q_i)XX^\top(w_i - w^q_i)^\top,
\end{align}
where $XX^\top$ happens to be the Hessian approximation derived from the Fisher information matrix with respect to the layer-wise target, sharing the same structure as Equation (\ref{eq:sb}). In the following discussion, we refer to this objective as \textit{L-full}. Similarly, we can also take the diagonal approximation here, by assuming that $X$ has an expectation close to zero. We use \textit{L-diag} to refer to this approximated objective:
\begin{align}
    & \mathop{\arg\min}\limits\limits_{w^q_i}\  (w_i - w^q_i)\ \mathrm{diag}(XX^\top)\ (w_i - w^q_i)^\top \\
    =\ & \mathop{\arg\min}\limits\limits_{w^q} \sum_{j=1}^m \left\|x_{j}\right\|^2 \cdot(w_{ij}-w^q_{ij})^2. \label{eq:ld}
\end{align}

\paragraph{Sensitivity-based quantization} For the $i$-th row of $W$ and $Y$, the relationship $y_i = w_i X$ holds. Consequently, the gradient relationship can be expressed as $g_{w_i} = g_{y_i} X^\top$. Using this result, Equation (\ref{eq:sb}) with Fisher approximation can be rewritten as:
\begin{align}
    & \mathop{\arg\min}\limits\limits_{w^q_i}\ (w_i - w^q_i)g_{w_i}^\top g_{w_i}(w_i - w^q_i)^\top \\
    =\ & \mathop{\arg\min}\limits\limits_{w^q} \left( (w_i - w^q_i)Xg_{y_i}^\top\right)^2.
\end{align}

This Equation shares the same form as Equation (\ref{eq:lf}) and will be referred to as \textit{S-full} in the following text. Similarly, Equation (\ref{eq:sb-diag}) can also be expressed in a form that includes the output gradient and input:
\begin{equation}
\mathop{\arg\min}\limits\limits_{w^q} \sum_{j=1}^m \left(x_j \cdot g_{y_i}^\top\right)^2 \cdot(w_{ij}-w^q_{ij})^2.
\end{equation}
This form is structurally identical to Equation (\ref{eq:ld}) and will be denoted as \textit{S-diag}. 

To conclude, the above transformation highlights the similarities and core differences between layer-wise and sensitivity-based quantization, allowing us to intuitively evaluate their effectiveness and make selections, as detailed in Section \ref{sec:obj} under the context of our SKIM method.

\section{Methodology}
\label{sec:method}
The entire process of our SKIM algorithm is detailed in Figure \ref{fig:process}. It begins by allocating different bits to each channel using a greedy algorithm, based on their quantization errors. If channel $i$ is allocated $b_i$ bits, $2^{b_i}$ centroids will be generated for it. Next, weighted K-means clustering is applied to each channel to calculate the centroids and labels, using the allocated bits as a constraint. Finally, we incorporate the scaling vector and train it through an iterative optimization strategy. We keep the labels fixed to enable gradient-based optimization, which means the vector only adjusts centroids. The dequantized weights can be recovered from the final labels and centroids, along with the scaling vector. Full algorithm of our SKIM method is illustrated in Appendix \ref{sec:full}. In the following subsections, we will explain our mixed precision and scaling vector techniques, as well as the principles behind our objective selection, in detail.

\subsection{Objective Selection}
\label{sec:obj}
As discussed, our SKIM framework consists of three main steps: the greedy algorithm, weighted K-Means clustering, and the scaling vector. A key question arises regarding how optimization objectives should be selected for these steps. Specifically, should the same objective be used consistently across all steps, or is cross-objective optimization feasible? Moreover, under different computational scenarios, which objective is the most effective, and how should it be chosen?

These questions can be addressed through our previous reformulation, from which we can conclude that although the layer-wise and sensitivity-based objectives differ, they can ultimately be transformed into either the \textit{full} or \textit{diag} forms. In these forms, both objectives exhibit a similar structure, ensuring their synchronization towards the final goal and allowing for cross-objective optimization in our work. Related experimental results are detailed in Section \ref{sec:iter}. The only difference lies in whether $g_{y_i}$ is introduced to serve as a guide for the subsequent model architecture.

Therefore, we can intuitively assess the effectiveness of each objective in the following order, from best to worst: \textit{S-full}, \textit{L-full}, \textit{S-diag}, and \textit{L-diag}. This is because the \textit{S} form incorporates gradient information as a guide, while the \textit{diag} form takes an aggressive diagonal approximation. The analysis aligns with our experimental results, which will be discussed in more detail in Section \ref{sec:eff}.

However, due to the interdependence among the elements of the gradient and the input, we cannot compute the four objectives using directly recorded expectations of $g$ or $X$. This limitation prevents us from adopting the most effective \textit{S-full} objective, as recording the corresponding $E(g^\top g)$ for each row requires quadratic memory complexity which is impractical for LLMs.

Consequently, in any scenario requiring a complete error calculation, such as the proposed mixed precision and scaling vector methods, we adopt the \textit{L-full} form. In contrast, for scenarios involving element-wise sums, such as the weighted K-means clustering, we adopt the \textit{S-diag} form. This will be the main principle for our objective selection.

\subsection{Mixed Precision}
\label{sec:mix}
\begin{figure}[b!]
  \centering
  \includegraphics[width=\columnwidth]{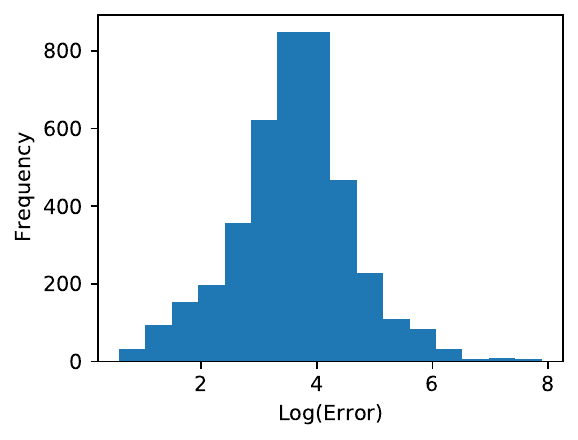}
  \vspace{-20pt}
  \caption{Histogram of the channel-wise quantization error for the $self\_attn.q\_proj$ in the second layer of Llama-7B. Errors vary significantly and exhibits a long-tail distribution on the larger side.}
  \label{fig:err}
\end{figure}

\begin{figure}[t!]
  \centering
  \includegraphics[width=\columnwidth]{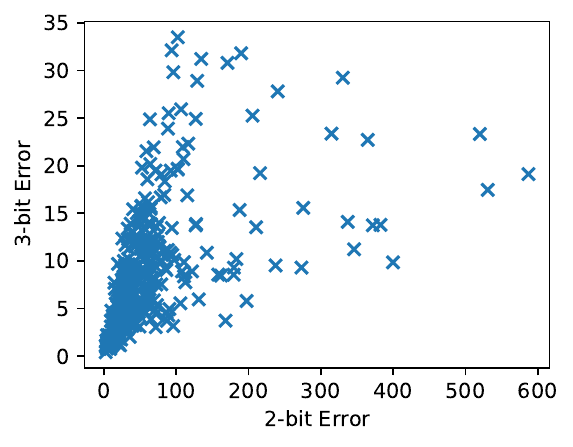}
  \vspace{-20pt}
  \caption{Error variation of the $self\_attn.q\_proj$ in the first layer of Llama-7B. We randomly sampled 10\% of the total rows for clearer visualization, with each point representing one. The horizontal axis indicates the quantization error when using 2 bits, while the vertical axis shows the error after increasing the bit level from 2 to 3. It is important to note that after the increase, same previous quantization error does not imply a similar post-increase error, and larger error does not lead to a larger result as well.}
  \label{fig:errrel}
\end{figure}

As previously mentioned, the quantization error of a weight matrix $W$ is the sum of that of its individual rows. However, each row, represent a channel, soften exhibits distinct data distributions and quantization errors, as illustrated in Figure \ref{fig:err}. Additionally, Figure \ref{fig:errrel} demonstrates that the quantization error after a one-bit increase is also unpredictable, even knowing the error at current bit. These observations indicate that applying the same bit-width for quantizing all rows results in an disproportionate allocation of resources. Motivated by these insights, we propose the adaptive channel-wise \textbf{Mixed Precision} technique, which aims to solve:
\begin{equation}
    \mathop{\arg\min}\limits\limits_{b_1,b_2,\dots,b_n} \sum_{i=1}^m Err(w_i, b_i),\ s.t.\ \frac{1}{n}\sum_{i=1}^mb_i \le \hat{b},
\end{equation}

where $b_i$ represents the bit allocated for channel $i$, $\hat{b}$ is the total bit constraint, and $Err(\cdot)$ function refers to the \textit{L-full} form of error between original $w_i$ and $b_i$ bit quantized weight, following our previous analysis. This formulation identifies the mixed precision issue as a bit-constrained sum minimization problem, which can be viewed as a variation of the knapsack problem \cite{martello1987algorithms, CACCHIANI2022105692} and precisely solved using a dynamic programming algorithm. However, the quadratic computational complexity of this algorithm renders it infeasible for scalability in LLMs. Therefore, we opt for a faster greedy algorithm to approximate the optimal solution. The greedy algorithm can yield a solution sufficiently tight to the optimal one while only requires a time complexity of $O(n\log n)$. Similar greedy algorithms has been widely adopted in various topics related to language models \cite{chen2021actnn, liu2022gact}, demonstrating its effectiveness. \cref{alg:bit} describes the specific steps in detail.
\begin{algorithm}[h!]
    \caption{Algorithm for Bit Allocation}
    \label{alg:bit}
    \begin{algorithmic}
       \INPUT Error Matrix $E$, Minimum Available Bit $b_\text{min}$, Maximum Available Bit $b_\text{max}$, Row Number $n$, Bit $bit$
       \OUTPUT Bit Allocation $b$
       \STATE Initialize $b = \{b_\text{min}\}_n\in \mathcal{R}^n$\;
       \REPEAT
       \STATE $i = \arg\max_{0 \le i < n} \{E_{i,b_i+1}-E_{i,b_i}|\ b_i<b_\text{max}\}$
       \STATE $b_i = b_i+1$
       \UNTIL{$\sum_{i=1}^nb_i \ge n\cdot bit$}
       \STATE \textbf{return} $b$
    \end{algorithmic}
\end{algorithm}

Additionally, our works includes efforts to enhance efficiency in mixed precision. Regarding computation for K-means clusterings, we develop an efficient framework utilizing parallel execution and shared memory. Furthermore, to accelerate the packing and unpacking phases with mixed precision, we gather all rows with the same bit and process them together, leveraging the massive computational capabilities of modern GPUs. You can found details about the time efficiency of our method in Appendix \ref{sec:qtime}.

\subsection{Scaling Vector}
\label{sec:scale}
Inspired by previous works \cite{xiao2023smoothquant}, we recognized that a scaling vector could capably regularize variations across columns, complementing our mixed-precision method. This operation can be formally expressed as: \[
    W^q = Quant(W * \alpha^{-1}) * \alpha,
\]
where $W$ is the original weight, $*$ represents the element-wise multiplication with broadcasting, and $\alpha^{-1}$ indicates the element-wise reciprocal of the scaling vector.

However, unlike previous works, a significant challenge in our scenario lies in determining the values of the scaling vector $\alpha$. Specifically, outliers play a less dominant role in weight distributions compared to activations, and under channel-wise quantization, the most intuitive approaches based on similarity measures cannot be directly applied. Empirical values suggested by SmoothQuant \cite{xiao2023smoothquant} and similar choices fail in our setting, leading to counterintuitive increases in perplexity. This highlights the need for a robust, automated approach to derive scaling values, rather than relying on manual assignment.

An intuitive solution is to make the scaling vector learnable and optimize it toward a reasonable target, similar to OmniQuant \cite{shao2023omniquant}, which employs gradient descent on block-wise error to train equivalent transformations, including scaling and shifting. However, the presence of a non-differentiable grouping operator in K-means clustering prevents the direct application of gradient descent to solve for $\alpha$. To tackle this challenge, we propose a novel strategy called iterative optimization, which optimizes grouping and scaling separately. By keeping the grouping results fixed during training, this approach enables gradient computation for the scaling vector. Algorithm \ref{alg:train} provides a detailed explanation of the strategy. Note that the loss calculation adopts the \textit{L-full} form of error, as discussed in our previous analysis. Additionally, when mixed precision is enabled, the $Kmeans(\cdot)$ function uses the allocation results from Algorithm \ref{alg:bit}, quantizing each channel with different precisions.

Considering the joint optimization problem of the clustering results and scaling vector, iterative training can be actually viewed as a cross-objective coordinate descent algorithm \cite{wright2015coordinate}. This raises a natural question: do different objectives interfere with the optimization of the algorithm? The answer is negative, and this conclusion can be supported both analytically and empirically. Our previous analysis shows the similarity between objective, ensuring their synchronization towards the final goal. And experiments also align with this analysis, as evidenced by results in Section \ref{sec:iter}. Experiment results also demonstrate that a single iteration is sufficient for the algorithm to produce adequate results, giving us the confidence to limit the maximum number of iterations to one for the sake of time efficiency.

\begin{algorithm}[t!]
    \caption{Iterative Optimization}
    \label{alg:train}
    \begin{algorithmic}
        \INPUT Weight $W$, Iteration $I$ = 1
        \OUTPUT Labels $L$, Centroids $c$, Scaling Vector $\alpha$
        \STATE $n, m = W.shape$
        \STATE Initialize $\boldsymbol{\alpha} = \mathbf{1}_{1\times m}\in \mathcal{R}^{1\times m}$
        \FOR{$i=1$ {\bfseries to} $I$}
            \STATE $\tilde{W} = W*\alpha^{-1}$ \hfill \textit{\ \ // element-wise multiplication}
            \STATE $L = \mathbf{0}_{n\times m} \in \mathcal{R}^{n\times m}$\;
            \FOR{$i=1$ {\bfseries to} $n$}
                \STATE \textit{// optimize on grouping}
                \STATE $L_{i,:}\leftarrow Kmeans(\tilde{W}_{i, :}).labels$ 
            \ENDFOR
            \REPEAT 
                \STATE $c = calc\_centroids(W * \boldsymbol{\alpha}^{-1}, L)$\;
                \STATE $W^q = replace(c, L) * \boldsymbol{\alpha}$\;
                \STATE $loss = Err(W, W^q)$\;
                \STATE $loss.backward()$\; \hfill \textit{// optimize on scaling}
            \UNTIL{Converged}
        \ENDFOR
        \STATE \textbf{return} $L,\ c,\ \alpha$
    \end{algorithmic}
\end{algorithm}

\section{Experiments}
\begin{table*}[t!]
    \caption{\textbf{Quantization Result of LLaMA models. } We report perplexity of quantized LLaMA-7B, LLaMA-13B and LLaMA-30B in this table. Note that SqueezeLLM did not provide an kernel implementation for 2-bit setting, so we merge their official code with our functions and report the reproduced results. Perplexity of OPT models and comparison with other baselines can be found in Appendix \ref{sec:addexp}.}
    \label{table:ppl-llama}
    \vskip 0.15in
    \begin{center}
        \begin{tabularx}{\textwidth}{l|YYY|YYY|YYY|YYY}
            \toprule
            \multirow{3}{*}{LLaMA-7B} & \multicolumn{3}{c|}{4 bit} & \multicolumn{3}{c|}{3.x bit} & \multicolumn{3}{c|}{3 bit} & \multicolumn{3}{c}{2.x bit} \\
            \cline{2-13}
            & \multirow{2}{*}{Bit} & \multicolumn{2}{c|}{PPL} & \multirow{2}{*}{Bit} & \multicolumn{2}{c|}{PPL} & \multirow{2}{*}{Bit} & \multicolumn{2}{c|}{PPL} & \multirow{2}{*}{Bit} & \multicolumn{2}{c}{PPL} \\
            & & Wiki & C4 & & Wiki & C4 & & Wiki & C4 & & Wiki & C4 \\
            \midrule
            FP16 & - & 5.68 & 7.08 & - & 5.68 & 7.08 & - & 5.68 & 7.08 & - & 5.68 & 7.08 \\
            SqueezeLLM & 4 & \textbf{5.79} & 7.21 & 3.24 & 6.13 & 7.56 & 3 & 6.32 & 7.75 & 2.23 & 11.32 & 15.69 \\
            OmniQuant & 4 & 5.86 & 7.34 & 3.24 & 6.15 & 7.75 & 3 & 6.48 & 8.19 & 2.25 & 9.72 & 12.79 \\
            \rowcolor{blue!10} SKIM & 4 & \textbf{5.79} & \textbf{7.20} & 3.2 & \textbf{6.07} & \textbf{7.52} & 3 & \textbf{6.21} & \textbf{7.68} & 2.25 & \textbf{8.99} & \textbf{11.00} \\
            \bottomrule
        \end{tabularx}
        \vskip 0.15in
        \begin{tabularx}{\textwidth}{l|YYY|YYY|YYY|YYY}
            \toprule
            \multirow{3}{*}{LLaMA-13B} & \multicolumn{3}{c|}{4 bit} & \multicolumn{3}{c|}{3.x bit} & \multicolumn{3}{c|}{3 bit} & \multicolumn{3}{c}{2.x bit} \\
            \cline{2-13}
            & \multirow{2}{*}{Bit} & \multicolumn{2}{c|}{PPL} & \multirow{2}{*}{Bit} & \multicolumn{2}{c|}{PPL} & \multirow{2}{*}{Bit} & \multicolumn{2}{c|}{PPL} & \multirow{2}{*}{Bit} & \multicolumn{2}{c}{PPL} \\
            & & Wiki & C4 & & Wiki & C4 & & Wiki & C4 & & Wiki & C4 \\
            \midrule
            FP16 & - & 5.09 & 6.61 & - & 5.09 & 6.61 & - & 5.09 & 6.61 & - & 5.09 & 6.61 \\
            SqueezeLLM & 4 & 5.18 & 6.71 & 3.25 & 5.45 & 6.92 & 3 & 5.60 & 7.08 & 2.23 & 8.74 & 12.57 \\
            OmniQuant & 4 & 5.21 & 6.76 & 3.25 & 5.44 & 7.05 & 3 & 5.68 & 7.32 & 2.24 & 7.93 & 10.76 \\
            \rowcolor{blue!10} SKIM & 4 & \textbf{5.17} & \textbf{6.70} & 3.2 & \textbf{5.42} & \textbf{6.92} & 3 & \textbf{5.52} & \textbf{7.04} & 2.25 & \textbf{7.40} & \textbf{9.22}\\
            \bottomrule
        \end{tabularx}
        \vskip 0.15in
        \begin{tabularx}{\textwidth}{l|YYY|YYY|YYY|YYY}
            \toprule
            \multirow{3}{*}{LLaMA-30B} & \multicolumn{3}{c|}{4 bit} & \multicolumn{3}{c|}{3.x bit} & \multicolumn{3}{c|}{3 bit} & \multicolumn{3}{c}{2.x bit} \\
            \cline{2-13}
            & \multirow{2}{*}{Bit} & \multicolumn{2}{c|}{PPL} & \multirow{2}{*}{Bit} & \multicolumn{2}{c|}{PPL} & \multirow{2}{*}{Bit} & \multicolumn{2}{c|}{PPL} & \multirow{2}{*}{Bit} & \multicolumn{2}{c}{PPL} \\
            & & Wiki & C4 & & Wiki & C4 & & Wiki & C4 & & Wiki & C4 \\
            \midrule
            FP16 & - & 4.10 & 5.98 & - & 4.10 & 5.98 & - & 4.10 & 5.98 & - & 4.10 & 5.98 \\
            SqueezeLLM & 4 & 4.22 & 6.06 & 3.25 & \textbf{4.44} & 6.23 & 3 & 4.66 & 6.37 & 2.22 & - & - \\
            OmniQuant & 4 & 4.25 & 6.11 & 3.25 & 4.56 & 6.37 & 3 & 4.74 & 6.57 & 2.24 & 6.59 & 9.36 \\
            \rowcolor{blue!10} SKIM & 4 & \textbf{4.20} & \textbf{6.05} & 3.2 & 4.46 & \textbf{6.22} & 3 & \textbf{4.57} & \textbf{6.31} & 2.25 & \textbf{5.80} & \textbf{7.49} \\
            \bottomrule
        \end{tabularx}        
    \end{center}
    \vskip -0.1in
\end{table*}

\subsection{Setups} 
\paragraph{Quantization Details} We evaluate our method within the context of post-training and weight-only quantization. The default setting includes INT4 and INT3, as well as INT3 and INT2 with extra memory usage. Note that we have set the maximum available bit to 4 in order to maintain high memory efficiency. Consequently mixed precision is disabled under the INT4 setting. And to optimize the scaling vector, we utilize the Adam \cite{kingma2014adam} optimizer with a learning rate of 0.01, a decrease rate of 0.5 every 40 steps and a maximum number of iterations of 120.

\paragraph{Baselines} We primarily compare our method against two baselines: SqueezeLLM \cite{kim2023squeezellm}, OmniQuant \cite{shao2023omniquant}. SqueezeLLM provides state-of-the-art performance under both INT4 and INT3 settings, while OmniQuant offers greater flexibility and performs better in the INT2 setting. Since most existing works offer limited options for bit levels, we ensure fairness by aligning our chosen bit levels with those that are more widely adopted and only comparing under similar conditions. Comparison with other baselines, including DecoupleQ \cite{guo2024decoupleq} and ABQ-LLM \cite{zeng2024abq}, is in Appendix \ref{sec:cmpother}, as they only reported results on limited datasets or bit levels.

\paragraph{Models and Datasets} We test across various model families and sizes, including LLaMA models(7B-30B) \cite{touvron2023llama} and OPT (2.7M-6.7B) models \cite{zhang2022opt}, to assess the generalizability of our method. We emphasize the results on LLaMA models in the main text due to their superior performance compared to other open-source LLMs and their widespread adoption. Comprehensive results for OPT models can be found in Appendix \ref{sec:pplopt}. Regarding datasets, we primarily utilize WikiText2 \cite{merity2016pointer} and C4 \cite{raffel2020exploring} for evaluation, along with 100 samples from the C4 dataset for calibration.

\vspace{-2pt}
\paragraph{Evalutation} We use the perplexity of language generation experiments as one of our primary evaluation metrics. Specifically, we report the perplexity on both the WikiText2 \cite{merity2016pointer} and C4 \cite{raffel2020exploring} datasets. Since our calibration dataset is generated from C4, the perplexity on WikiText2 represents a zero-shot evaluation, while the perplexity on C4 can be considered a few-shot scenario. We do not present the specific average bits under integer settings, as the differences between methods are negligible and it is clearer to compare them using unified bit levels. Additionally, we assess accuracy on the MMLU \cite{hendrycks2020measuring} benchmark, which spans a diverse range of domains. This assessment is conducted under both zero-shot and five-shot scenarios, aiming to evaluate the problem-solving capability of our quantized model.

\subsection{Perplexity Results}
The results of our SKIM method applied to the LLaMA models from 7B to 30B are presented in Table \ref{table:ppl-llama}. We conducted a comparative evaluation of our method against other ones across various bit levels, consistently finding that SKIM outperforms the alternatives. These findings highlight the versatility of SKIM, demonstrating its adaptability to a wide range of configurations. Notably, with 3-bit quantization, our method achieves a significant reduction in perplexity, narrowing the performance gap between full precision and 3-bit quantized models by 18.5\% on LLaMA-7B, 15.7\% on LLaMA-13B and 16.1\% on LLaMA-20B. Furthermore, our 3.2-bit model even surpasses others that operates at a slightly higher bit level, showcasing the effectiveness of our approach. These substantial reductions are also evident in other models, as detailed in Appendix \ref{sec:addexp}.

\begin{table*}[t!]
    \centering
    \caption{Comparison of averaged MMLU accuracy on LLaMA-7B models. Since SqueezeLLM does not provide accuracy for LLaMA models, we benchmarked the model generated by their official code to get the results.}
    \label{tab:mmlu}
    \vskip 0.15in
    \begin{tabularx}{\textwidth}{l|c|YYYYY|YYYYY}
        \toprule
        \multirow{3}{*}{LLaMA-7B} & \multirow{3}{*}{Bit} & \multicolumn{5}{c|}{0-shot} & \multicolumn{5}{c}{5-shot} \\
        & & \multirow{2}{*}{Hum.} & \multirow{2}{*}{STEM} & Social & \multirow{2}{*}{Other} & \multirow{2}{*}{Avg.} & \multirow{2}{*}{Hum.} & \multirow{2}{*}{STEM} & Social & \multirow{2}{*}{Other} & \multirow{2}{*}{Avg.} \\
        & & & & Sci. & & & & & Sci. & & \\
        \midrule
        Baseline & 16 & 33.0\% & 27.4\% & 32.4\% & 37.3\% & 32.7\% & 30.6\% & 34.0\% & 38.4\% & 38.3\% & 35.1\% \\
        \midrule
        SqueezeLLM & 3 & 27.2\% & 28.0\% & 25.2\% & 27.5\% & 27.2\% & 30.2\% & 27.8\% & 31.8\% & 35.3\% & 31.2\% \\
        \rowcolor{blue!10} SKIM & 3 & 29.4\% & 26.2\% & 26.9\% & 30.1\% & \textbf{28.3\%} & 31.8\% & 31.6\% & 34.4\% & 35.5\% & \textbf{33.2\%}  \\
        \midrule
        SqueezeLLM & 3.24 & 29.4\% & 27.0\% & 30.1\% & 33.1\% & \textbf{29.9\%} & 28.4\% & 30.4\% & 32.0\% & 34.5\% & 31.3\% \\
        \rowcolor{blue!10} SKIM & 3.2 & 30.4\% & 27.7\% & 28.5\% & 31.8\% & 29.7\% & 32.3\% & 31.2\% & 34.0\% & 36.6\% & \textbf{33.2\%} \\
        \bottomrule
    \end{tabularx}
\end{table*}

\subsection{MMLU Benchmarking} 
We evaluate our quantized models on the MMLU benchmark, providing domain-specific and average accuracy for both zero-shot and five-shot settings. Since SqueezeLLM does not offer benchmarking results for LLaMA models, we benchmark the quantized model generated by their official code. As shown in Table \ref{tab:mmlu}, our SKIM method improves the performance of the quantized LLaMA-7B in both zero-shot and five-shot scenarios, as evidenced by increased or comparable average accuracy. The evaluation spans multiple domains, with the calibration dataset having no direct relation to them. Therefore, the performance improvement strongly demonstrates method's ability to retain the knowledging capabilities of the model.

\subsection{Ablation Study}
\subsubsection{Effectiveness of Optimization Objectives}
\label{sec:eff}
As mentioned earlier, different forms and contents of the optimization objectives can lead to varying levels of effectiveness. Here, effectiveness refers to the extent to which the final loss or perplexity is positively influenced by optimizing a specific target. In practice, we selected two scenarios to test the objectives: the sampled weight for K-means clustering and the loss calculation for training the scaling vector. In the K-means clustering scenario, both mixed precision and scaling vector are disabled, while in the scaling vector scenario we take \textit{S-diag} as weights and mixed precision is disabled. We use 3-bit quantization and the final perplexity to illustrate their effectiveness. Detailed results can be found in Table \ref{tab:eff}, and the findings align perfectly with our theoretical analysis. We opted not to test the \textit{S-full} form due to its quadratic complexity, which renders it impractical.
\begin{table}[htb!]
    \centering
    \begin{tabular}{c|c}
        \toprule
        Objective & Perplexity \\
        \midrule
        \textit{L-full} & - \\
        \textit{S-diag} & 6.33 \\
        \textit{L-diag} & 6.36 \\
        \bottomrule
    \end{tabular}
    \hspace{0.02\columnwidth}
    \begin{tabular}{c|c}
        \toprule
        Objective & Perplexity \\
        \midrule
        \textit{L-full} & 6.24 \\
        \textit{S-diag} & 6.27 \\
        \textit{L-diag} & 6.29 \\
        \bottomrule  
    \end{tabular}
    \caption{Comparison of effectiveness between different objectives. On the left, we present the perplexity results for weighted K-means clustering, excluding the results of \textit{L-full} due to its non-conformity to the element-wise sum. On the right, we display the perplexity results for the scaling vector scenario.}
    \label{tab:eff}
    \vspace{-10pt}
\end{table}

\begin{figure}[b!]
    \centering
    \includegraphics[width=\columnwidth]{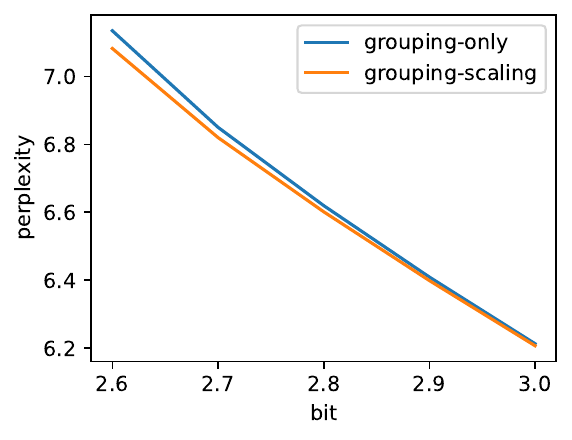}
    \vspace{-20pt}
    \caption{Perplexity variation after enabling scaling vector. Perplexity consistently decreases when additional optimization on the scaling vector is applied.}
    \label{fig:iter}
    \vspace{-8pt}
\end{figure}

\subsubsection{Iterative Optimization}
\label{sec:iter}
Separating the optimization and assigning different objectives does not cause fluctuations in perplexity, as illustrated in Figure \ref{fig:iter}. When scaling is disabled, we obtain the grouping-only curve; enabling scaling produces the grouping-scaling curve. The consistent drop in perplexity after enabling the vector validates the synchronization between the targets. Regarding the number of iterations, we empirically find that the perplexity reduction brought by extra iterations is much more modest compared to the significant decrease observed after the first iteration between grouping and scaling. Additionally, increasing the number of iterations can sometimes lead to overfitting, which causes slight fluctuations in perplexity. Therefore, we are confident to set the maximum number of iterations to one.



\subsection{Memory Efficiency}
Following the approach outlined in SqueezeLLM \cite{kim2023squeezellm}, we evaluated the Peak Memory Usage of our SKIM method when generating 64 tokens, as detailed in Table \ref{tab:mem}. Our method achieved lower peak memory usage even at the same bit level due to the utilization of lower precision for cluster centroids. Additionally, our 3.2-bit quantized model demonstrates greater memory savings compared to the 3.24-bit model given by SqueezeLLM, while also excelling in performance. Beyond memory efficiency, our approach stands out for its ability to break the fixed bit grid. Given a specific GPU capacity constraint, users can select the maximum bit level to fully exploit the machine. For more detailed information, please refer to Appendix \ref{sec:mem}.

\begin{table}[ht]
    \centering
    \begin{tabularx}{0.9\columnwidth}{l|Y|Y}
        \toprule
        LLaMA-7B\  & 3bit & 3.xbit \\
        \midrule
        FP16 & \multicolumn{2}{c}{12.72GB} \\
        \midrule
        Squeeze & 3.01GB & 3.26GB    \\
        SKIM & 2.98GB & 3.13GB   \\
        \bottomrule
    \end{tabularx}
    \\ \ \\ \ \\
    \begin{tabularx}{0.9\columnwidth}{l|Y|Y}
        \toprule
        LLaMA-13B & 3bit & 3.xbit \\
        \midrule
        FP16 & \multicolumn{2}{c}{24.63GB} \\
        \midrule
        Squeeze & 5.45GB & 5.88GB    \\
        SKIM & 5.41GB & 5.69GB   \\
        \bottomrule
    \end{tabularx}
    \caption{Memory efficiency with LLaMA-7B and LLaMA-13B.}
    \label{tab:mem}
\end{table}

\section{Conclusion}
We propose \textbf{S}caled \textbf{K}-means clustering w\textbf{I}th \textbf{M}ixed Precision (SKIM), an effective posting-training and weight-only quantization method. Building on previous non-uniform quantization methods, SKIM further incorporate two novel techniques: Adaptive Mixed Precision and Trainable Scaling Vector. Our method is evaluated across a wide range of models, tasks, and bit levels, consistently outperforming previous state-of-the-art methods. Its memory efficiency and ability to break the fixed bit grid facilitate the deployment of large language models.

\nocite{langley00}
\bibliographystyle{icml2024}
\bibliography{paper}

\newpage
\appendix
\onecolumn
\section{Full Algorithm}
\label{sec:full}
The full algorithm for our SKIM method is illustrated in Algorithm \ref{alg:full}. Note that we provide the simplified version with number of iterations equal to one. Our method includes three main steps: 1. compute the bit allocation (line 2-11); 2. apply channel-wise K-Means Clustering (line 12-15); 3. train the scaling vector (line 16-22). Additionally, error pre-recording (line 2-10) only needs to be executed once for each model, and all KMeans functions are accelerated using multi-processing and shared memory.

\begin{algorithm}[h!]
    \caption{Overall Algorithm for SKIM}
    \label{alg:full}
    \begin{algorithmic}[1]
        \INPUT Weight $W$, Gradient Square $G$, Bit $bit$, Minimum Available Bit $b_\text{min}$, Maximum Available Bit $b_\text{max}$
        \OUTPUT Labels $L$, Centroids $c$, Scaling Vector $\alpha$
        \STATE $n, m = W.shape$ 
        \STATE $E = [\ 0\ ]_{n\times(b_\text{max} - b_\text{min} + 1)} \in R^{n\times(b_\text{max} - b_\text{min} + 1)}$  \hfill \textit{// Error Matrix}
        \FOR{$i=1$ {\bfseries to} $n$}
            \FOR{$\hat{b}=b_\text{min}$ {\bfseries to} $b_\text{max}$}
            \STATE $result\leftarrow Kmeans(W_{i, :},\ weights =G_{i,:},\ n\_centriods=2^{\hat{b}})$ \hfill \textit{// apply Kmeans($\cdot$) with actual configuration}
            \STATE $l, c \leftarrow result.labels,\ result.centroids$ 
            \STATE $w^q = replace(c, l)$ \hfill \textit{// replace labels with corresponding centroids}
            \STATE $L_{i,\hat{b}}=Err(W_{i,:},w^q)$ \hfill \textit{// pre-record quantization error with Equation \ref{eq:lf}}
            \ENDFOR 
        \ENDFOR
        \STATE $b = alloc\_bit(E,\ b_\text{min},\ b_\text{max},\ n, \ bit) \in \mathbb{R}^n$  \hfill \textit{// compute bit allocation with Algorithm \ref{alg:bit}}
        \STATE $L = \mathbf{0}_{n\times m} \in \mathcal{R}^{n\times m}$\;
        \FOR{$i=1$ {\bfseries to} $n$}
            \STATE $L_{i,:}\leftarrow Kmeans(W_{i, :},\ weights =G_{i,:},\ n\_centriods=2^{b_i}).labels$ \hfill \textit{// optimize on grouping towards Equation \ref{eq:sb-diag}}
        \ENDFOR
        \STATE Initialize $\boldsymbol{\alpha} = \mathbf{1}_{1\times m}\in \mathcal{R}^{1\times m}$
        \REPEAT 
            \STATE $c = calc\_centroids(W * \boldsymbol{\alpha}^{-1}, L)$\;
            \STATE $W^q = replace(c, L) * \boldsymbol{\alpha}$\; \hfill \textit{// replace labels with centroids and unscale to reconstruct the weight}
            \STATE $loss = Err(W, W^q)$\; \hfill \textit{// calculate quantization error with Equation \ref{eq:lf}}
            \STATE $loss.backward()$\; \hfill \textit{// optimize on scaling}
        \UNTIL{Converged}
        \STATE \textbf{return} $L$, $c$, $\alpha$
    \end{algorithmic}
\end{algorithm}

\section{Additional Experiment Result}
\label{sec:addexp}
\subsection{Perplexity Evaluation on Other Models}
\label{sec:pplopt}
Table \ref{table:ppl-other} contains all results on OPT models. Our method continues to outperforms others, showcasing its generalizability. However, on OPT models, mixed precision would greatly reduces the quantization error yet slightly increase the final perplexity. For example, on the \textit{up\_proj} layer mixed precision reduces the quantization error by more than 50\% but increases the perplexity. We attribute this phenomenon to overfitting on the calibration data, and address it by disabling mixed precision under integer bit levels while initializing the bit allocation with flooring to the specified bits under non-integer levels. Consequently, even with the slight increase in perplexity, our method still provides better or comparable results.
\begin{table*}[ht!]
    \caption{\textbf{Quantization Result of OPT models}}
    \label{table:ppl-other}
    \vskip 0.15in
    \begin{center}
        \begin{tabularx}{\textwidth}{l|YYY|YYY|YYY|YYY}
            \toprule
            \multirow{3}{*}{OPT-2.7B} & \multicolumn{3}{c|}{4 bit} & \multicolumn{3}{c|}{3.x bit} & \multicolumn{3}{c|}{3 bit} & \multicolumn{3}{c}{2.x bit} \\
            \cline{2-13}
            & \multirow{2}{*}{Bit} & \multicolumn{2}{c|}{PPL} & \multirow{2}{*}{Bit} & \multicolumn{2}{c|}{PPL} & \multirow{2}{*}{Bit} & \multicolumn{2}{c|}{PPL} & \multirow{2}{*}{Bit} & \multicolumn{2}{c}{PPL} \\
            & & Wiki & C4 & & Wiki & C4 & & Wiki & C4 & & Wiki & C4 \\
            \midrule
            FP16 & - & 12.47 & 13.17 & - & 12.47 & 13.17 & - & 12.47 & 13.17 & - & 12.47 & 13.17 \\
            SqueezeLLM & 4.07 & 12.80 & 13.38 & 3.25 & 13.43 & 
            \textbf{13.88} & 3 & 13.85 & 14.45 & - & - & - \\
            OmniQuant & 4 & 12.76 & 13.58 & 3.24 & 13.18 & 14.15 & 3 & 13.80 & 14.93 & 2.25 & \textbf{18.13} & 21.11 \\
            \rowcolor{blue!10} SKIM & 4 & \textbf{12.72} & \textbf{13.35} & 3.2 & \textbf{13.34} & 13.92 & 3 & \textbf{13.66} & \textbf{14.21} & 2.25 & 19.79 & \textbf{19.96} \\
            \bottomrule
        \end{tabularx}
        \vskip 0.15in
        \begin{tabularx}{\textwidth}{l|YYY|YYY|YYY|YYY}
            \toprule
            \multirow{3}{*}{OPT-6.7B} & \multicolumn{3}{c|}{4 bit} & \multicolumn{3}{c|}{3.x bit} & \multicolumn{3}{c|}{3 bit} & \multicolumn{3}{c}{2.x bit} \\
            \cline{2-13}
            & \multirow{2}{*}{Bit} & \multicolumn{2}{c|}{PPL} & \multirow{2}{*}{Bit} & \multicolumn{2}{c|}{PPL} & \multirow{2}{*}{Bit} & \multicolumn{2}{c|}{PPL} & \multirow{2}{*}{Bit} & \multicolumn{2}{c}{PPL} \\
            & & Wiki & C4 & & Wiki & C4 & & Wiki & C4 & & Wiki & C4 \\
            \midrule
            FP16 & - & 10.12 & 11.20 & - & 10.12 & 11.20 & - & 10.12 & 11.20 & - & 10.12 & 11.20 \\
            SqueezeLLM & 4 & 11.03 & 11.85 & 3.26 & 11.31 & \textbf{12.18} & 3 & 11.70 & 12.44 & - & - & - \\
            OmniQuant & 4 & 11.03 & 11.97 & 3.25 & \textbf{11.27} & 12.31 & 3 & 11.65 & 12.78 & 2.25 & \textbf{14.43} & 16.67 \\
            \rowcolor{blue!10} SKIM & 4 & \textbf{11.02} & \textbf{11.84} & 3.2 & \textbf{11.27} & 12.20 & 3 & \textbf{11.46} & \textbf{12.39} & 2.25 & 14.79 & \textbf{16.01} \\
            \bottomrule
        \end{tabularx}

    \end{center}
    \vskip -0.1in
\end{table*}

\begin{table*}[bt!]
    \caption{\textbf{Comparation with DecoupleQ. Perplexity on WikiText2 is reported in this table.}}
    \label{table:ppl-cmpdq}
    \vskip 0.15in
    \centering
    \begin{tabularx}{0.6\textwidth}{l|YY|YY|YY}
            \toprule
            \multirow{2}{*}{Method} & \multicolumn{2}{c|}{4 bit} &  \multicolumn{2}{c|}{3 bit} & \multicolumn{2}{c}{2.x bit} \\
            \cline{2-3} \cline{4-5} \cline{6-7}
            & Bits & PPL($\downarrow$) & Bits & PPL($\downarrow$) & Bits & PPL($\downarrow$) \\
            \midrule
            LLaMA-7B & - & 5.68 & - & 5.68 & - & 5.68 \\
            DecoupleQ & 4 & 5.85 & 3 & 6.38 & 2.25 & 8.65 \\
            \rowcolor{blue!10} SKIM & 4 & \textbf{5.79} & 3 & \textbf{6.70} & 2.30 & \textbf{8.64} \\
            \midrule
            LLaMA-13B & - & 5.09 & - & 5.09 & - & 5.09 \\
            DecoupleQ & 4 & 5.21 & 3 & 5.60 & 2.25 & 7.25 \\
            \rowcolor{blue!10} SKIM & 4 & \textbf{5.17} & 3 & \textbf{5.52} & 2.30 & \textbf{7.11} \\
            \midrule
            LLaMA-30B & - & 4.10 & - & 4.10 & - & 4.10 \\
            DecoupleQ & 4 & 4.24 & 3 & 4.67 & 2.25 & 6.04 \\
            \rowcolor{blue!10} SKIM & 4 & \textbf{4.20} & 3 & \textbf{4.57} & 2.30 & \textbf{5.66} \\
            \bottomrule
        \end{tabularx}
\end{table*}
\subsection{Comparation with other baselines}
\label{sec:cmpother}
DecoupleQ\cite{guo2024decoupleq} decouples model parameters into integer and floating point parts, achieving state-of-the-art performance in certain low-bit configurations. Table \ref{table:ppl-cmpabq} compares our SKIM method with DecoupleQ, and C4 results in excluded as DecoupleQ does not provide corresponding results. SKIM consistently outperforms DecoupleQ across most settings by a significant margin, highlighting its effectiveness. When operating at INT2 precision with additional memory, SKIM provides comparable performance to DecoupleQ at the same bit level. However, with just a slight increase in memory- specifically, 0.05 bits-our method surpasses DecoupleQ by a considerable margin. We present results with this minimal extra memory to illustrate the flexibility of our approach.

ABQ-LLM\cite{zeng2024abq} proposes a block-wise distribution correlation and compensation schema for Post-Training Quantization, demonstrating strong performance at lower bit-widths as well. We report perplexity results on LLaMA-7B and LLaMA-13B models for INT4 and INT3 settings, as these overlap with our experiment data. As shown in Table \ref{table:ppl-cmpabq}, SKIM outperforms ABQ-LLM in both the INT4 and INT3 configurations.

\begin{table*}[thb!]
    \caption{\textbf{Comparation with ABQ-LLM. Perplexity on both WikiText2 and C4 dataset is reported.}}
    \label{table:ppl-cmpabq}
    \vskip 0.15in
    \centering
    \begin{tabularx}{0.42\textwidth}{l|YY|YY}
        \toprule
        \multirow{2}{*}{LLaMA-7B} & \multicolumn{2}{c|}{4 bit} & \multicolumn{2}{c}{3 bit} \\
        & Wiki & C4 & Wiki & C4  \\
        \midrule
        FP16 & 5.68 & 7.08 & 5.68 & 7.08 \\
        ABQ-LLM & 5.83 & 7.29 & 6.29 & 8.01 \\
        \rowcolor{blue!10} SKIM & \textbf{5.79} & \textbf{7.20} & \textbf{6.21} & \textbf{7.68} \\
        \bottomrule
    \end{tabularx}
    \hspace{0.03\textwidth}
    \begin{tabularx}{0.42\textwidth}{l|YY|YY}
        \toprule
        \multirow{2}{*}{LLaMA-13B} & \multicolumn{2}{c|}{4 bit} & \multicolumn{2}{c}{3 bit} \\
        & Wiki & C4 & Wiki & C4  \\
        \midrule
        FP16 & 5.09 & 6.61 & 5.09 & 6.61 \\
        ABQ-LLM & 5.19 & 6.75 & 5.56 & 7.24 \\
        \rowcolor{blue!10} SKIM & \textbf{5.17} & \textbf{6.70} & \textbf{5.52} & \textbf{7.04} \\
        \bottomrule
    \end{tabularx}
\end{table*}

\section{Memory Usage With Varying Bit Levels}
\label{sec:mem}
\begin{figure}[ht!]
    \centering
    \includegraphics[width=0.6\textwidth]{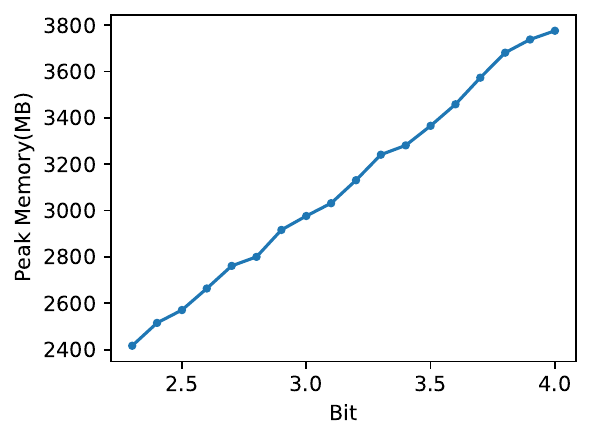}
    \vspace{-10pt}
    \caption{The actual peak memory usage of LLaMA-7B when generating 64 tokens. }
    \label{fig:bitmem}
\end{figure}

Figure \ref{fig:bitmem} illustrates the correlation between memory usage and bit levels. The peak memory usage increases linearly with the bit level, which facilitates the selection of an appropriate level to meet specific memory capacity requirements and to maximize the model's performance under a specific machine.

\section{Quantization Cost}
\subsection{Memory Demands}
As a post-training quantization method, SKIM inherits the positive characteristics of memory savings. Although it involves training, this process is layer-wise, and the scaling vector is the only trainable parameter. As a result, the memory requirements during the quantization phase are significantly lower than those during inference. For instance, quantizing LLaMA-7B requires peak memory usage of less than 8GB, which is well within the memory capacity of most modern GPUs.

\subsection{Quantization Time}
\label{sec:qtime}
In terms of K-means clustering, we leverage parallel execution and shared memory to enhance efficiency. Speed for both error matrix pre-recording and K-means quantization benefits from these improvements. To record the errors of LLaMA-7B using our framework, the process takes less than half an hour on dual AMD EPYC processors, which feature a total of 128 cores and 256 threads. And once the errors are recorded, they can be utilized for quantization at any specified bit level without needing to repeat the process. For K-means quantization, we can process approximately 8000 rows per second under the same machine conditions, while a transformer block in LLaMA-7B only contains 42496 rows.

Regarding the packing and unpacking phases, by avoiding sparse matrices and consolidating all rows with same bit level, the packing phase is significantly faster than that of SqueezeLLM. It takes about one minute to pack our quantized LLaMA-7B, whereas SqueezeLLM typically exceeds five minutes.

Overall, the entire process for quantizing LLaMA-7B takes around one hour with dual AMD EPYC processors and an RTX 3090 GPU. Compared to OmniQuant, our method is more efficient as we break the block-wise training to layer-wise one. And compared to SqueezeLLM, although the time consumption is slightly higher due to the computations involved in bit allocation and scaling vector training, it still remains comparable, ensuring high efficiency.


\end{document}